\documentclass[12pt]{article}
\usepackage{scicite}

\usepackage{times}

\usepackage{amsmath}
\usepackage{url}
\usepackage{amsfonts}
\usepackage{amssymb}
\usepackage{graphicx}
\usepackage{xspace}

\usepackage{booktabs}

\usepackage[font=footnotesize]{caption}
\usepackage{changepage}

\newenvironment{sciabstract}{%
\begin{quote} \bf}
{\end{quote}}

\newcommand{\et}{et al.\xspace}

\title{Uncovering the Social Interaction in Swarm Intelligence with Network Science} 

\author
{M. Oliveira,${}^{1\ast}$ D. Pinheiro,${}^{2}$ M. Macedo,${}^{3}$ C. Bastos-Filho,${}^{4}$  R. Menezes${}^{3}$\\
\\
\normalsize{${}^{1}$Computational Social Science, GESIS--Leibniz Institute for the Social Sciences, Germany}\\
\normalsize{${}^{2}$Department of Internal Medicine, University of California, Davis, Sacramento, USA}\\
\normalsize{${}^{3}$Department of Computer Science, University of Exeter, Devon, UK}\\
\normalsize{${}^{4}$Polytechnic School of Pernambuco, University of Pernambuco, Recife, Brazil}\\
\\
\normalsize{$^\ast$To whom correspondence should be addressed; E-mail:  moliveira@tuta.io.}
}

\date{}

\usepackage{xcolor}
\usepackage{soul}
\definecolor{reddish}{HTML}{FBB4AE}
\definecolor{blueish}{HTML}{B3CDE3}
\definecolor{magentish}{HTML}{FF00AA}
\definecolor{greenish}{HTML}{a1d99b}

\begin{document} 




\maketitle




\begin{sciabstract}
Swarm intelligence is the collective behavior emerging in systems with locally interacting components. Because of their self-organization capabilities, swarm-based systems show essential properties for handling real-world problems such as robustness, scalability, and flexibility. Yet, we do not know {\itshape why} swarm-based algorithms work well and neither we can compare the different approaches in the literature. The lack of a common framework capable of characterizing these several swarm-based algorithms, transcending their particularities, has led to a stream of publications inspired by different aspects of nature without a systematic comparison over existing approaches. Here, we address this gap by introducing a network-based framework---the interaction network---to examine computational swarm-based systems via the optics of the {\itshape social dynamics} of such interaction network; a clear example of network science being applied to bring further clarity to a complicated field within artificial intelligence. We discuss the social interactions of four well-known swarm-based algorithms and provide an in-depth case study of the Particle Swarm Optimization. The interaction network enables researchers to study {\itshape swarm algorithms as systems}, removing the algorithm particularities from the analyses while focusing on the structure of the social interactions. 
\end{sciabstract}



\section*{Introduction}
\label{sec:introduction}
Swarm intelligence refers  to  the  global  order that emerges from simple social components interacting among themselves\cite{Bonabeau1999,Kaufmann1993,Vicsek2001,Kennedy2001,Engelbrecht2006}. In the past three decades, swarm intelligence has inspired many algorithmic models (i.e., computational swarm intelligence), allowing us to understand social phenomena and to solve real-world problems\cite{Engelbrecht2006}. The field of computational intelligence has witnessed the development of various swarm-based techniques that share the principle of social interactions while having different natural inspirations such as ants\cite{Dorigo1999}, fishes\cite{bastos2008novel}, fireflies\cite{yang2010firefly}, birds\cite{Kennedy95}, cats \cite{Chu:Cat:2006}, to name a few. Though researchers have studied such techniques in detail, the lack of general approaches to assess these systems prevents us from uncovering what makes them intelligent and understanding the differences between techniques beyond their inspirations.

Much research has been devoted to understand and improve these bio-inspired algorithms\cite{Kennedy2001,Engelbrecht2006,Engelbrecht2007}. In the literature, researchers often examine the techniques from the perspective of their natural inspirations. For instance, in some flocking models that mimic bird flocks, the velocities of individuals are usually used to understand the system behavior\cite{Engelbrecht2007}. In these systems, both the lack or excess of spatial coordination among individuals generally leads to poor performance in solving problems. In the case of foraging-based models inspired by ant colonies, many studies attempt to understand the performance of these models by examining the pheromone that agents deposit on the environment\cite{Dorigo2004}. This usual approach of analyzing models via their inspiration has helped to improve algorithms by building new procedures\cite{SunNew2004,dong2017supervised}.

These analyses, however, are confined to specific niches that have their metaphor (e.g., ants following pheromone, birds searching for food, fireflies trying to synchronize) and jargon (e.g., pheromone, velocity, fish weight). The broad variety of natural inspirations makes it challenging to find interchangeable concepts between swarm intelligence techniques\cite{Sorensen2015}. The absence of niche-free analyses restricts the findings of a model to its own narrowed sub-field. Such myopia leads us to miss the underlying mechanisms driving a system to the undesired states that new techniques (or procedures) endlessly try to avoid. In this scenario, we need agnostic quantitative approaches to analyze computational swarm intelligence in a general manner and thus provide the means to understand and improve algorithms in whatever niche.

\begin{figure}[b!]
\centering
\includegraphics[width=4.5in]{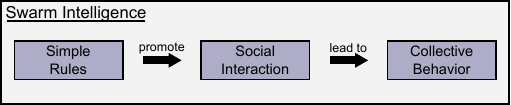}
\caption{
Social interactions at the meso level are still overlooked by researches who often devote considerable efforts to understand how changing the simple rules at the micro level (e.g., procedures, equations) directly affects the collective behavior of the system at the macro level. In fact, these micro-level rules create the conditions to social interaction at the {\itshape meso level} which in turn enables the necessary swarm dynamics to solve complex problems at the macro level. 
} \label{fig:flowchart}
\end{figure}

The field lacks general methodologies to analyze swarms because of the absence of a generic framework to examine their main similarity: the social interactions  (see~\figurename~\ref{fig:flowchart}). Indeed, the concept of {\it social interaction} is fundamental in swarm intelligence; it refers to the exchange of information through diverse mechanisms\cite{Bonabeau1999,Kennedy2001}. In this definition, social interactions are not only the mere exchange of information between peers but also have the potential to change individuals\cite{Kennedy2001}. The sophisticated behavior emerging from social interactions enables the system to adjust itself to solve problems\cite{Kennedy2001}. In swarm intelligence techniques, individuals process information and locally interact among themselves, spreading knowledge within the swarm, which results in the emergent system ability. In this sense, examining the social mechanisms is fundamental to understand intelligence in these systems. This general perspective also helps us to assess swarms with different natural inspirations. Instead of relying on the complete understanding of the micro-level properties (e.g., velocity, pheromone, weight), we can assess the swarm via the structure and dynamics of the social interactions\cite{Bonabeau1999}.

Notably, the field of Network Science has shown that every complex system can be represented as a network encoding the interactions between the components of the system and that the understanding of the structure of this network is {\it sine qua non} for learning the behavior of the system itself\cite{barabasi2011network}. Network Science advocates that the complex systems comprehension can be reached by observing the structure and dynamics of their underlying networks\cite{Strogatz2001,barabasi2011network}. Though the idea of using networks as frameworks for understanding complex phenomena dates back to Moreno's use of sociograms in the 1940s\cite{moreno1946sociogram}, it has been popularized by two seminal papers from Watts and Strogatz\cite{Watts_Collective}, and Barab{\'a}si and Albert\cite{barabasia99} in the late 1990s. Recent works in the field have demonstrated that even small variations in fundamental structural properties, such as the degree distribution, can significantly influence the behavior of the system described by the network.

Here we propose a network-based framework---the interaction network---for the general assessment and comparison of swarm intelligence techniques. With this framework, we study swarm algorithms as systems, removing the niche-specific particularities from the analyses. In the following sections, we start by describing the importance of understanding swarm-based algorithms, and by explaining the definition of the interaction network. We show how the interaction network can be defined for four well-known swarm-based algorithms from two distinct categories as proposed by Mamei \et\cite{Mamei2006}, namely flocking and foraging. Then, we demonstrate a complete case study using the concept of flocking and show the relationship between the interaction network and the swarm behavior. 

\section*{Understanding Swarm Systems}

In the field of Computational Swarm Intelligence, scholars often analyze algorithms via their performance on given problems. In many cases, innovation means the development of novel algorithms that are capable of achieving improved results on a set of benchmark functions. These improvements, however, tend to arise without much explanation. Also, researchers often use jargons in both the justification for novel algorithms and the description of their improvements\cite{Sorensen2015}. This black-box approach not only hinders general interpretability of results but also sidetracks us from the underlying mechanisms driving the swarm intelligence in these systems. The case occurs because of the lack of a unifying view of swarm-based algorithms.  Though some efforts have been made to understand swarm systems from a general perspective, they tend to be qualitative in nature. 

\begin{figure}[b!]
\centering

\includegraphics[width=4in]{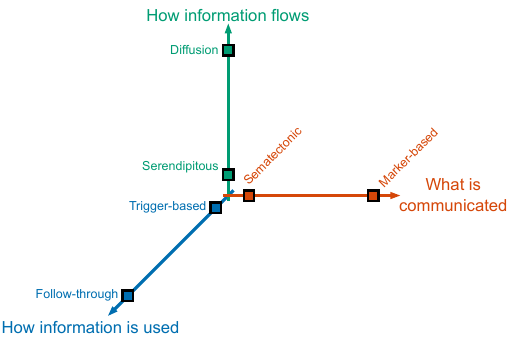}
\caption{%
Three dimensions of information processing define the self-organized mechanism in a swarm system: how information flows, how information is used, and the definition of information. The {\it diffusion} flow occurs when individuals passively receive information that other individuals spread in the environment whereas {\it serendipitous} flow occurs when individuals need to actively search for information left in the environment by other individuals. When using information in a \textit{trigger-based} system, individuals act in the environment by performing specific, mostly one-off, actions,  while in a \textit{follow-through}, they are guided by what they find, and the action can be more long-lasting. The {\it marker-based} information is explicitly defined for interaction purposes~(e.g., pheromone), while individuals implicitly share {\it sematectonic} information as the current state of the population (figure adapted from\cite{Mamei2006}).
}
\label{fig:dimensions}
\end{figure}

The general perspective for swarm-based systems proposed by Mamei \et is that of a system processing information\cite{Mamei2006}. From this perspective, the way a swarm handles information defines its underlying self-organization mechanism. We can describe a system using three aspects of information: (i) the definition of information, (ii) how individuals use information, and (iii) how information flows within the system (see \figurename~\ref{fig:dimensions}). This approach {\it classifies} swarm systems but fails to examine them quantitatively. 

In fact, the literature has various approaches to classify swarm systems\cite{Parpinelli2011,Duan2015,Chu2018} and metaheuristics in general\cite{Gendreau2005,Fernandez-Marquez2013,Fong2016}. These efforts are essential to organize the field. They are the necessary initial steps to understand current and new algorithms. Still, the absence of quantitative approaches prevents us from characterizing the particularities of methods and quantifying their differences. 

\begin{figure*}[t]
\centering
\begin{adjustwidth}{-.5in}{0in} 
\includegraphics[width=7.5in]{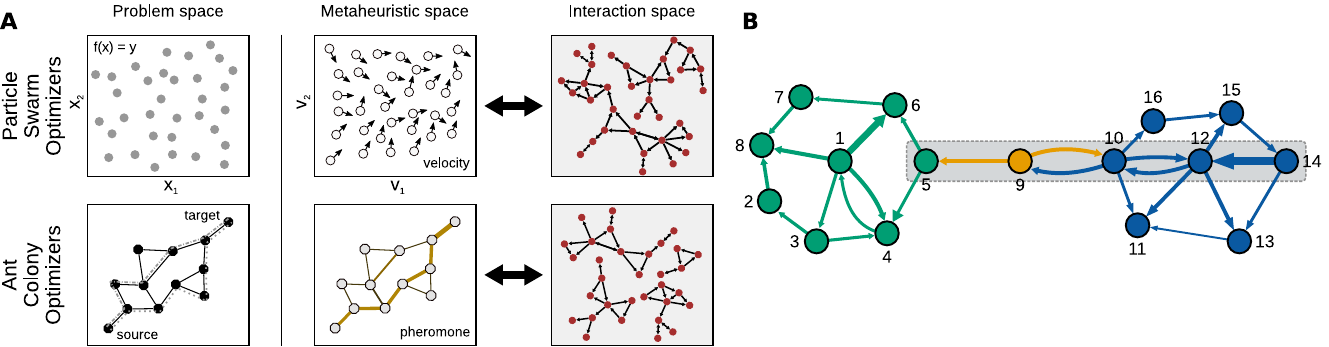}
  \end{adjustwidth}
\caption{
{Different viewpoints of swarm systems.} We highlight that the interaction space is a general way of viewing these systems. (\textbf{A}) Models of swarm intelligence are commonly used to solve continuous (top) and discrete (bottom) problems that can be represented in the \textit{problem space} (left). Each model incorporates metaphor-specific aspects such as velocities (first row) and pheromone (second row) in PSO and ACO, respectively, that can be represented in the \textit{metaheuristic space} (middle). However, regardless of the problem and the metaphor, the actions of individuals are interdependent and depend on the social interactions. The use of the interaction network allows us to represent these social interactions in the \textit{interaction space} (right), enabling a unified assessment of metaheuristics. (\textbf{B}) A general illustration of an interaction network for swarm systems where each node is an individual in the population, and each link represents the direction and the extent of the individuals' influence. Each color depicts a distinct sub-network to which members are highly integrated and tightly connected when compared to nodes outside the sub-network.} \label{fig-explanation}
\end{figure*}

In some cases, researchers measure the swarm  diversity to understand the technique\cite{Chu2018}. This diversity is often the diversity of the candidate solutions when solving a given problem\cite{Engelbrecht2007,Dorigo2004,Krink2002,ShiEberhart2008,Olorunda2008,Shi2009}. With such an approach, however, we focus on the final outcome of the swarm dynamics, neglecting the underlying mechanism leading to these dynamics. We lack a framework enabling us to examine the system from an intermediate perspective.

\subsection*{The Social Interaction in Swarm Systems}
\label{sec:mechanisms}

The dynamics of swarm-based systems depend on social interaction. The system lacks coordination without enough interaction among the individuals and loses adaptability with the excess\cite{chate2014viewpoint}. In such systems, the local rules promote or undermine the level of interaction within the swarm (\figurename~\ref{fig:flowchart}). In this sense, the social interactions are halfway between the micro and macro behavior of the system. The network emerging from these complex interactions is a natural universal meso-level perspective of swarms.

Previous research has used the network paradigm to examine the emergent behavior in social animals and its underlying mechanisms\cite{Fewell2003,Lusseau2003,Strandburg-Peshkin2013,Rosenthal2015}. Some developments in the computational intelligence field have also taken advantage of networks\cite{Giacobini2006,Payne2009,4983016,Dorronsoro2012,pso_complenet2012,Liu2014,Metlicka2015}. They have been used to understand swarm systems\cite{Whitacre2008,Huepe2011} and their respective collective behaviors such as flocking\cite{complenet2013,Oliveira2014,Oliveira2015,Oliveira2016,Oliveira2017a,Janostik2016b,Pluhacek2016,Du2016} and foraging\cite{Metlicka2015,Kromer2016}. In this regard, Oliveira \et proposed one of the first approaches to examine interactions within the swarm in the Particle Swarm Optimization\cite{complenet2013}. Yet, these preliminary efforts have focused on specific techniques, missing the fact that social interaction is the common feature driving swarm intelligence.

In this work, we argue that the social dynamics in swarm-based algorithms should be more analyzed and explored to provide insights into the {\it dynamic network} behind the rules and inspirations, which may lead to a possible meta-classification of the systems---a meta-classification based on the system dynamics instead of the natural inspiration of the system. In the following section, we define the interaction network and use the categorization proposed by Mamei \et to elaborate on the plausibility of describing and employing the interaction network to assess models of swarm intelligence inspired by different mechanisms of self-organization. Through different mechanisms, an interaction network can be built to characterize the system over a common space: the interaction space. In this space, the interaction network becomes a general framework that allows for the unified assessment of swarm intelligence models with distinct inspirations.

\section*{The Network of Social Interaction}
We propose to examine the social interaction within a swarm as a way to assess the behavior of swarm intelligence systems. Here we develop the concept of {\it interaction network} to represent the interdependencies of the actions of the individuals. For a given swarm system, the interaction network $\mathbf{I}$ consists of nodes that represent its individuals and edges $\textbf{I}_{ij}$ that indicate the extent to which individual $i$ influences the action of the individual~$j$. As social interactions are dynamic and so the swarm, we use $\mathbf{I}(t)$ to describe the influence that individuals exert on each other at time $t$. 

The interaction network is a representation of the swarm and the result of the rules that define the swarm system. Though these rules are bio-inspired, the network $\mathbf{I}$ belongs to the \textit{interaction space} $\mathcal{I}$~(see~\figurename~\ref{fig-explanation}A). This is an agnostic space exempt from the particularities of the swarm algorithm or problems being solved by the algorithm.  Note that both the algorithm (i.e., rules) and problem modify the social dynamics within the system and have an impact on $\mathbf{I}$. Yet, when we look at algorithms from this general framework, we have the potential to assess different algorithms that are, at their surface, completely distinct (i.e., inspired by distinct natural phenomena).

The network structure---at both global- and individual-levels---enables us to analyze different aspects of the swarm, and aspects across different swarm intelligence approaches. For instance, \figurename~\ref{fig-explanation}B depicts a conceptual interaction network for swarm systems. At the individual level, the network positions occupied by individuals indicate the types of interdependencies that were created by the swarm and the influence individuals may exert on one another. The individuals with a high degree centrality (e.g., individuals $1$ and $12$) typically exert stronger influence when compared to other individuals. Similarly, individuals that connect different groups (e.g., individual $9$) act as bridges between groups of individuals and control the cascade of influence between sub-networks. Thus, some individuals in a swarm system can develop important roles as bridges and hubs. Lastly, at a global level, the interaction network indicates the extent of local and global exploration by providing the relationship between natural niches formed by individuals (e.g., green and blue sub-networks).

To analyze a swarm using the interaction network, we need to learn the rules and mechanisms that allow individuals to influence the action of each other within the system. We use the dimensions described in \figurename~\ref{fig:dimensions} to guide our understanding of algorithms and thus to define their networks. For a given algorithm, we have to identify how an individual uses information after it received information from other individuals. It is clear that we need first to characterize what information is in the system then describe how information exchange influences individuals. Table \ref{tab:mechanisms} describes how the interaction network can be constructed for each category proposed by Mamei \et\cite{Mamei2006}.

\begin{table}[b!]
\begin{adjustwidth}{-.5in}{0in} 

\caption{Social interactions of the eight swarm intelligence categories as organized by Mamei \et\cite{Mamei2006}.} 
\small
   
\label{tab:mechanisms}
\begin{center}
\tabcolsep=0.17cm
\begin{tabular}{lp{1.8cm}p{2.9cm}p{1.5cm}p{6.4cm}p{1.47cm}}
    \toprule    Name          & Inspiration     & Heuristic      & Comm.    & Social interaction     & Example \\
    
    \midrule    Quorum        &  bacteria            
    & decision-making
    & 
  Direct  & \raggedright Triggered based on the passively perceived current arrangement of their local environment             & BSO\cite{ali2012coordinated}       \\
    
    \midrule    Embryogenesis & cells\cite{lawrence1992making}            & 
    self-assembling
    & 
    Direct & \raggedright Triggered based on the passively perceived markers of social interaction diffused in the environment & 
    GSO\cite{krishnanand2005detection}       \\
    \midrule    Molding       &  \raggedright slime mold amoebas\cite{Schmickl2007} & \raggedright swarm coordinated aggregation
    & 
    Direct & \raggedright Passively perceived markers of social interaction they diffuse in the environment             & Graduate Routing\cite{Poor2001}\\
    
    \midrule    Flocking      & birds\cite{Kennedy2001} & 
    \raggedright swarm coordinated motion
    & 
    Direct & \raggedright Continuously self-propel based on the current arrangement of their passively perceived environment                                  &PSO\cite{Kennedy95}    
    \\
    \midrule    Brood sorting & worker ants \cite{Bonabeau1999}   &
    sorting of items
    & 
    Indirect & \raggedright Triggered based on the actively found current arrangement of their local environment                &     SMOA\cite{monismith2008slime}  
    \\
    
    \midrule    Nest Building & termites\cite{Dynamic2014}          & 
    \raggedright coordinated construction of complex structures 
    &    
    Indirect & \raggedright Triggered based on the actively found markers of social interaction left in the environment & MBO\cite{abbass2001monogenous}       \\
    \midrule    Foraging      & ants\cite{Bonabeau1999} & 
    \raggedright coordinated exploration of the environment 
    & 
    Indirect & \raggedright Continuously self-propel based on the actively found markers of social reinforcement left in the environment & ACO\cite{Dorigo1999}     \\
    \midrule    Web Weaving   & spiders\cite{Bourjot2003} & 
    \raggedright coordinated construction of complex structures
    & 
    Indirect &   Continuously self-propel based on the actively found current arrangement of their local environment                               & RIO\cite{havens2008roach}          \\
    \bottomrule
    \end{tabular}
    \end{center}
    \end{adjustwidth}

\end{table}

In the following subsection, we use the interaction network to compare four swarm-based algorithms from two distinct categories, namely flocking and foraging. In this brief analysis, we show that simple definitions of the interaction network enable us to recognize differences and similarities among these algorithms. Then, we provide a more detailed case study of PSO in which we study the dynamics of the swarm at the meso level of the social interactions. 

\subsection*{Modeling the Interaction Network}
\label{sub:algorithms}

\begin{figure*}[t]
  \centering
  \begin{adjustwidth}{-.5in}{0in} 
  \includegraphics[width=7.1in]{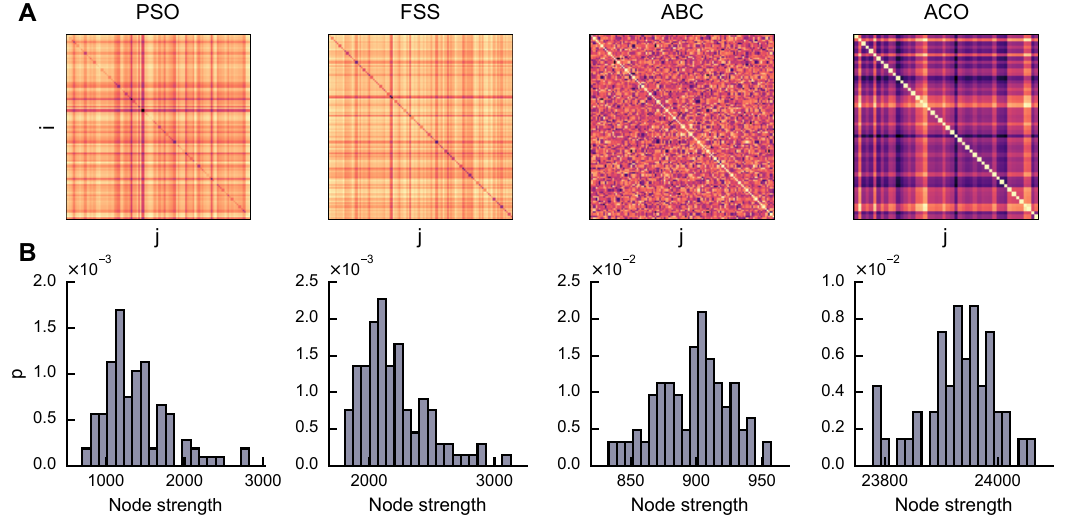}
  \end{adjustwidth}
  \caption{{The interaction network provides us the means to examine swarm-based algorithms from a general perspective.} Here we use simple definitions of the interaction network for four different algorithms: the Particle Swarm Optimization (PSO), the Fish School Search (FSS), the Artificial Bee Colony (ABC), and the Ant Colony Optimization (ACO). Though they have different bio-inspirations (i.e., bees, ants, birds, and fishes), we can analyze them in the same interaction space. For this, we build the interaction network $\mathbf{I}(t)$ for each of them based on the social operators in the algorithm. We sum up each matrix over time to analyze $\mathbf{I} = \sum_t \mathbf{I}(t)$. (\textbf{A}) After $700$ iterations, each algorithm shows distinct signatures. (\textbf{B}) The strength of a node (i.e., $\sum_j \mathbf{I}_{ij}$) tells us the influence of an individual on the population. Though PSO and FSS allow strong influencers to exist, ABC and ACO exhibit a well-behaved distribution of spreaders.}
    \label{fig:comparison_algorithms}
\end{figure*}

With the interaction network, we examine the social interaction in four bio-inspired optimization algorithms: the Artificial Bee Colony (ABC)\cite{karaboga2005idea}, the Ant Colony Optimization (ACO)\cite{Dorigo2004}, the Particle Swarm Optimization (PSO)\cite{Kennedy95}, and the Fish School Search (FSS)\cite{bastos2008novel}. These algorithms have distinct inspirations and are from distinct categories but share the same concept of social interaction. In this section, we avoid technicalities and save them to the next section, where we delve into a specific case study. Instead, here we discuss these algorithms from a high-level perspective, focusing on the social interactions. First, we describe how agents in the system influence each other; then, we build a network based on this description.

The core of these algorithms is that successful individuals are more likely to influence the population of the system. Though the definition of success depends on the algorithm, it generally relates to the quality of a solution. The agents navigate through a metaheuristic space relying on the information from other agents and the environment. We describe the interdependency in the actions (e.g., swimming, flying) of the individuals using the interaction network. For each algorithm, we build the network $\mathbf{I}(t)$ in that each link represents this interdependency between two individuals (i.e., nodes) at iteration $t$. 

In the case of the PSO algorithm, particles move towards the best particle (i.e., the most successful one) in their neighborhoods at each iteration $t$. To describe this influence with the interaction network $\mathbf{I}(t)$, we connect each agent $i$ to its best neighbor $j$ at each iteration $t$. Note that particles move using information from only \textit{one} individual in their neighborhood.

Fishes in the FSS algorithm use information from \textit{all} individuals in the fish school. Each fish contributes to the movement of the fish school based on its current displacement and its previous success. We describe this interdependency with a weighted network. We build a network in which the weight represents the proportional contribution of the individual $i$ to the collective movement of the individual $j$ at the iteration $t$. 

Note that both PSO and FSS are deterministic with regards to the interaction among individuals. The success of the agents \textit{determines} their interaction with other agents. In the case of the ABC algorithm, social interaction takes place only probabilistically. 

In the ABC algorithm, bees fly using information from bees that are selected based on a uniform distribution and a roulette wheel. The former enables an agent to influence any other agent regardless of success; whereas the latter tends to drive agents to interact with successful ones. We build the network $\mathbf{I}(t)$ based on this selection. Precisely, we create an edge (or increment the weight) between the agent $i$ to the agent $j$ every time the agent $i$ considered the position of agent $j$ to move at the iteration $t$.  

In the algorithms described above, individuals can communicate \textit{directly} with each other. Notably, the Ant Colony Optimization algorithm uses the concept of stigmergy in which the agents communicate only \textit{indirectly} via pheromones. In ACO, each ant moves across the environment following higher concentrations of pheromone and depositing pheromone at visited edges accordingly. In this case, we build a weighted interaction network by keeping track of the amount of pheromone deposited by each ant $i$ at each edge $e$. The influence $i$ exerts on $j$ is the cumulative amount of pheromones left by $i$ at the edges visited by $j$.

With these network definitions, we run an implementation of each algorithm and examine the social interaction in these systems. For each case, we analyze the swarm accounting for the whole execution, thus we sum up matrices over time as follows: $\mathbf{I} = \sum_t \mathbf{I}(t)$.  \figurename~\ref{fig:comparison_algorithms}A depicts the matrices for each network until $700$ iterations. From these heatmaps, we see similarities between PSO and FSS. Both algorithms enable the emergence of central individuals influencing the system. The probabilistic behavior of the ABC algorithm, however, prevents the rise of such patterns. The strength of the nodes (\figurename~\ref{fig:comparison_algorithms}B) helps us to study the social interactions in each algorithm. On the one hand, PSO and FSS allow the existence of strong spreaders; on the other hand, ABC and ACO have a more well-behaved distribution of spreaders.

The interaction network provides us the means to study these peculiarities in swarm intelligence algorithms from a general perspective---a necessary step towards understanding swarm intelligence. The framework also creates an opportunity to study meso-level dynamics, as we show in the next section. 

\section*{In-depth Case Study: Examining the Particle Swarm Optimization algorithm}
\label{sec:casestudy}

This section presents how to analyze the meso-level dynamics of a swarm-based algorithm using the interaction network. We selected the PSO algorithm because of its simplicity and wide use in several applications. 

Particle Swarm Optimization (PSO) is a population-based optimization method that relies on the interactions of individuals sharing the best positions they found during the search process\cite{Kennedy95}. The method---inspired by the social behavior in flocks of birds---consists of a population of simple reactive agents~(particles) that explore the search space by locally perceiving the environment and interacting among themselves to improve their solutions. 

In the standard definition of the PSO, each particle $i$ consists of four vectors in a $d$-dimensional search space: its current position $\vec{x_i}(t)$, its best position found so far $\vec{p}_{i}(t)$, its velocity $\vec{v}_{i}(t)$, and the best position found by its neighbors $\vec{n}_{i}(t)$\cite{Bratton2007}. The position of each particle represents a candidate solution to a $d$-dimensional continuous optimization problem, and the swarm moves through the problem search space seeking better solutions. To enable this capability, all particles change their positions, at each iteration $t$, according to their velocities $\vec{v}_{i}(t)$ which are updated based on the personal best position $\vec{p}_{i}(t)$ and the social best position $\vec{n}_{i}(t)$. Researchers use different ways to update the position of the particles, but the update equation generally aims to maintain the coherence of the particles through an inertia term and adjust the trajectory using cognitive and social information. In our study, we use the so-called constricted PSO\cite{clerc2002}.

The particles in the swarm only interact with a subset of the swarm. The swarm topology defines the infrastructure through which particles communicate and thus enables the particles to retrieve information from other particles (i.e., their neighbors). At each iteration $t$, each particle $i$ seeks for its best neighbor $n_i(t)$ in its neighborhood (i.e., the one with the best solution so far). The topology influences the social interaction within the swarm and has been shown to impact the swarm performance\cite{mendesPHD,Bratton2007}. Clerc proposed a somewhat different definition of swarm topology---the~so-called {\it graph of influence}---which explicitly includes the social information and presents directed edges\cite{clerc2010particle}. Regardless of definition, however, the swarm topology only refers to the structure for the potential exchange of information and neglects effective interaction among particles. 

In the particle-swarm context, exploration and exploitation refer to the ability of individuals to broadly explore the whole search space or focus on a particular area\cite{Kennedy2001}. An exploration--exploitation imbalance often leads to a poorly explored search space. To better understand this imbalance, researchers study the diversity and the dynamics of the swarm. The literature often focuses on the \textit{spatial} diversity\cite{ShiEberhart2008,Olorunda2008,Shi2009}, analyzing the outcomes of social interactions such as the positions or velocities of the particles in the search space. These approaches have succeeded in developing novel mechanisms to improve the performance of the algorithm. Yet, with these approaches, we fail to understand the \textit{underlying social interactions} driving the swarms to undesired states (e.g., lack of diversity, premature convergence) that new mechanisms try to avoid. 
  
Still, a few works have attempted to analyze the particles' interactions in order to examine the swarm behavior. Some of these efforts analyzed the impact of the infrastructure of the swarm communication on the swarm performance\cite{Kennedy2002,mendesPHD,Du2016}. Though these studies neglected the actual interactions between particles, they showed that bounding social interactions influences the swarm behavior. Oliveira \et were the first ones to examine the actual interactions among particles in order to assess the swarm\cite{complenet2013}. They proposed the analysis of the swarm using a network in which the nodes (particles) are connected if they share information in a given iteration and later extended the concept to capture historical information\cite{Oliveira2014,Oliveira2015,Oliveira2017a}. Later on, Pluhacek \et provided visualizations of the interactions in the swarm\cite{Pluhacek2016}.

In the next subsections, we define the {\it interaction network} $\mathbf{I}$ to assess the swarm using the methods developed by Oliveira \et\cite{Oliveira2016}. With this definition, we can uncover relationships between the swarm dynamics, the swarm performance, and the social interactions. 

\begin{figure*}[h!]
  \centering
  \begin{adjustwidth}{-.5in}{0in} 
   \includegraphics[width=7.5in]{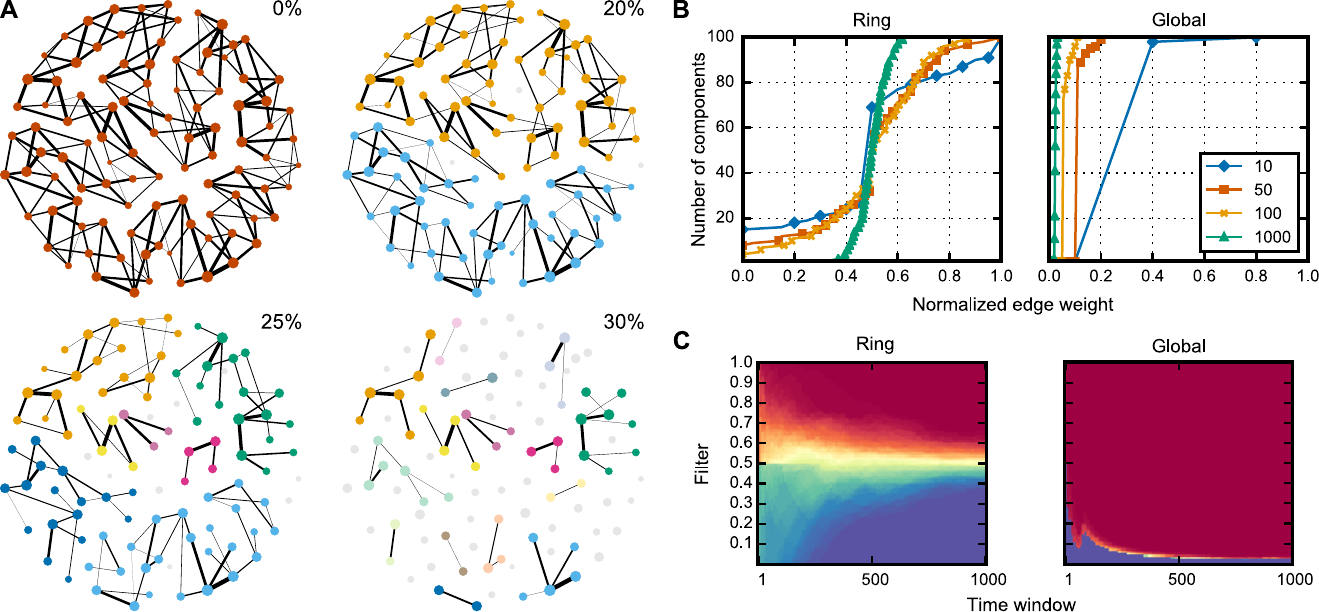} 
   \end{adjustwidth} 
  \caption{{The pace at which components emerge while edges are gradually removed from $\mathbf{I}$ is associated with the search mode of the swarm.} An exploration mode is characterized by a slow increase in the number of components due to the different information flows present in the swarm. The network, however, is rapidly destroyed in a swarm that depends only on a small set of individuals, a behavior related to an exploitation search mode. In PSO, (\textbf{A}) the weighted interaction network of a run with the swarm using a von~Neumann topology has edges removed based on their weight: below $20\%$ of the highest possible weight, $25\%$ and $30\%$. The colors represent components with more than one node. In this process, edges with the lowest weights are removed first. (\textbf{B})~The impact of the removal of the edges on the growth of the number of components depends on the structure of the swarm topology. The different colors/markers in the plot represent the time window $t_w$. The normalized weight is the weight value divided by $2t_w$, which is the highest possible weight in the network. The rapidly increasing in the number of components of the global topology leads to a type of behavior related to the exploitation search mode. In the ring topology, the number of components increases slowly, indicating the existence of sub-swarms searching more independently through the search space\cite{Oliveira2014}. In all cases, the swarm consists of 100 individuals. (\textbf{C})~Each topology leads to distinct interaction diversity that can be described by the number of components emerging~(color intensity) as edges are removed~(\textit{y}-axis) of the interaction network with different time windows (\textit{x}-axis). }
  \label{fig-panel3}
\end{figure*}

\subsection*{A Network for the Particle Swarm Optimization}
\label{sec:influence}
To examine a swarm system from the meso-level perspective of its social interactions, we need to build the interaction network to capture the structure and dynamics of the social influence exerted among individuals. In the case of the PSO algorithm, a social interaction occurs when a particle $i$ updates its position based on the position of a particle $j$. This happens when $j$ is the best neighbor of $i$ at a given iteration; that is, $n_i(t) = j$.

Here we use a simple (yet powerful) definition of interaction network~$\mathbf{I}(t)$ in which the weight of an edge $(i,j)$ is the number of times the particle $i$ was the best neighbor of the particle $j$ or vice-versa until the iteration $t$\cite{Oliveira2014}. We use a time window $t_w$ to control the recency of the analysis, thus the interaction network at iteration $t$ with window $t_w$ is defined as the following:
\begin{equation}
 \mathbf{I}_{ij}(t) = \sum_{t'=t-t_w+1}^{t}\bigg[\delta_{i,n_{j}(t')} + \delta_{j,n_{i}(t')}\bigg], 
  \label{eq:swarm_influence_graph_time}
\end{equation}
with $t \ge t_w \ge 1$ and where $\delta_{i,j}$ is Kronecker delta. In this definition, nodes (i.e., particles) are connected by an edge with weight equals to the number of times two particles shared information in at most $t_w$ iterations before the iteration $t$\cite{Oliveira2014}. The time window $t_w$ tunes the frequency--recency balance in the analysis. High $t_w$ makes the network dominated by most frequent interactions; low $t_w$ only includes most recent interactions and when $t_w = 1$ we have instantaneous interactions.

Note that the definition of an interaction network for a swarm system depends on the rules that promote social interaction in the system. Here we pinpointed that, in PSO, a social interaction between $i$ and $j$ occurs when the particle $i$ updates its velocity $\vec{v_{i}}$ using the position of a particle $j$. This definition of $\mathbf{I}$ is a simple one that includes only the occurrence of social interaction between particles. More complex definitions may include edge direction or other aspects of the algorithm, such as the social constant $c_2$ or the realizations of $\vec{r_2}$. Nevertheless, with this simple definition, we can already better understand the swarm\cite{complenet2013,Oliveira2014,Oliveira2015,Oliveira2016,Oliveira2017a}. Other swarm systems, however, have different rules and distinct forms of social interactions. 

\subsection*{Examining the Social Interaction with $\mathbf{I}$}

The formation of structures in the interaction network arises from the way information flows within the swarm, which, in turn, alters the dynamics of the swarm. The existence of well-connected nodes in $\mathbf{I}$ indicates frequent information flows in the swarm. The constant interaction among specific individuals leads to their respective nodes in the interaction network to be clustered. To capture these clusters, we can gradually remove the edges of $\mathbf{I}$ according to their weight; the components that emerge during this network destruction represent the information flows within the swarm~(see \figurename~\ref{fig-panel3}A). 

Note that the pace at which these components appear relates to the swarm dynamics. A slow increase suggests an exploration search mode in which individuals share information among distinct groups and thus create social interactions with various levels of tie strength. A rapid increase suggests, however, an exploitation search mode in which individuals interact with a few same sources and thus create a center of information with similar levels of tie strength.

With the definition in Eq.~\eqref{eq:swarm_influence_graph_time}, we can now examine the search mode in the PSO algorithm. For instance, we analyze $\mathbf{I}$ of swarms using different topology parameters---that are known to lead the swarm to behave differently---while solving the same problem. As shown in \figurename~\ref{fig-panel3}B, with the global topology, the particle swarm presents exploitation behavior, whereas the ring topology leads the system to explore different information sources. Note that this analysis differs from the typical analysis of the relationship between fitness and topology structure\cite{Kennedy2002,mendesPHD,6855840}. Here we focus on the way particles interact during the swarm search when using different structures: the communication topology affects the diversity of the \textit{interactions} in the swarm, generating different interaction networks.

To consider the swarm ability to maintain different frequent information flows, we can analyze the network destruction while varying $t_w$ to include frequency and recency in the analysis of the flows. \figurename~\ref{fig-panel3}C depicts the number of components that emerge when edges are removed from $\mathbf{I}$ with increasing time windows. The interaction network of a particle swarm with global topology seems to be destroyed at the same pace in both perspectives of frequency (i.e., high $t_w$) and recency (i.e., low $t_w$). The interactions of the particles within this topology promote a lack of diversity in the information flows in short and long terms. 

This diversity regards to the ability of the swarm to have a diverse flow of information---a perspective different from \textit{spatial} diversity in which $d$-dimensional properties of particles are compared to particular definitions of swarm center\cite{ShiEberhart2008}. Note that the lack of diversity in the information flow can decrease the spatial diversity in a swarm. The absence of multiple information flows leads to particles retrieving information from a few sources and drives particles to move towards the same region of the search space; lack of interaction diversity pushes individuals to the same direction.

To quantify interaction diversity, we measure the destruction pace of interaction networks with different time windows. For a given time window ${t_w}$, the area under the destruction curve $A_{t_w}$ can be seen as a measure of diversity in the information flow. High values of $A_{t_w}$ indicate fast destruction, whereas low values imply slow destruction.  Hence, we can define the interaction diversity ${\rm ID}$ (previously called communication diversity\cite{Oliveira2016}) as the mean diversity over a set of time windows $T$, as the following:
\begin{equation}
  {\rm ID}(t) = 1 - \dfrac{1}{|S||T|}\sum_{t'_w\in T}A_{t_w=t'_w}(t),
  \label{eq:cd}
\end{equation}
where $|S|$ is the number of particles in the swarm. 
Thus, swarms exhibiting high ${\rm ID}$ (i.e.,~low values for $A_{t_w}$) have the ability to have diverse information flows, while low values for ${\rm ID}$ imply swarms with only few information flows~(i.e., high value for $A_{t_w}$). The ideal set $T$ would be one taking into account all time windows (i.e., from $t_w=1$ until $t_w=t$). This procedure, however, can be computationally expensive given the vast number of possible time windows, so we use just a sample set of time windows. 

\subsubsection*{Experimental Design}
To investigate the extent to which the interaction diversity assesses the swarm at each iteration, we systematically examined the swarm using different topologies that lead the swarm to behave differently. We employ different connected $k$-regular graphs~(i.e., graphs in which nodes have $k$ links) as the swarm communication topology with $k$ ranging from $2$ to $100$. Here we consider a distinct group of four benchmark functions $F_2$, $F_6$, $F_{14}$, and $F_{19}$ from the CEC'2010 \cite{cec2010}. In all experiments, we set the number of dimensions to $1000$ and, when applicable, the degree of non-separability $m$ to $50$; and set the swarm size to $100$ particles. 

We analyze the relationship between ${\rm ID}$ and fitness improvement over time; thus we define fitness improvement $f_\Delta(t)$ at iteration $t$ as the speed at which the fitness $f_g(t)$ of the swarm changes between the two immediate iterations $t$ and $t-1$ as follows: $f_\Delta(t) = \frac{f_g(t) - f_g(t-1)}{f_g(t-1)}$, where $f_g(t)$ is the global best fitness of the swarm at iteration $t$.  In the simulations, we set as stopping criterion either a maximum number of iterations $t_{max}=10000$ is reached or the swarm has converged at iteration $t_s$. We define that a swarm converged at iteration $t_s$ if the global best fitness does not improve, that is, if $f_\Delta(t) < 10^{-5}$, until iteration $t_s+\delta$ with $\delta=500$. For each considered swarm topology, we run a PSO implementation $30$ times while measuring ${\rm ID}$ and $f_\Delta$ at each iteration in each execution.

\begin{figure}[b!]
\centering
\begin{adjustwidth}{-.5in}{0in} 
\includegraphics[width=6.9in]{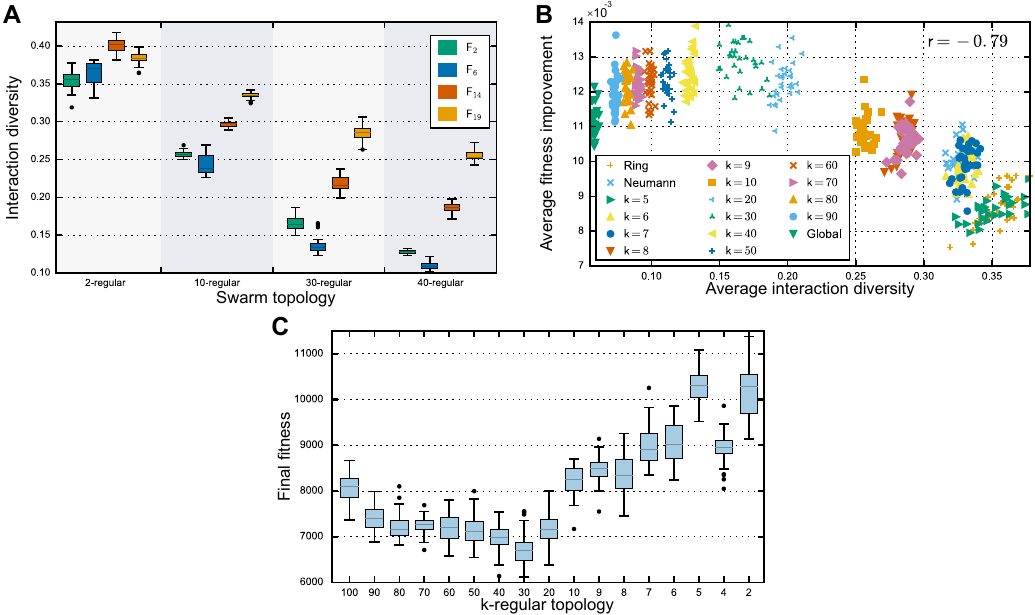}
\end{adjustwidth}
\caption{
{The interaction diversity, fitness improvement, final fitness, and $k$-topologies are associated in a non-trivial manner.} (\textbf{A}) Some benchmark functions appear to consistently present higher interaction diversity when compared to other regardless of the underlying communication topology. In the results for $F_6$ function, (\textbf{B}) although the correlation of $-0.79$ indicates a strong negative linear relationship between the average interaction diversity and the mean fitness improvement, one can easily see that they are associated in a non-monotonic way. (\textbf{C}) Similarly, the final quality of the fitness found by the swarm also presents a non-monotonic behavior regarding $k$-regular topologies and consequently, interaction diversity. }
\label{fig:cd_iteration}
\end{figure}

\subsubsection*{Results} 
We analyze the impact of the infrastructure of communication (i.e., topology) on the diversity of the information flows within a swarm. We found that $k$-regular topologies promote higher diversity as $k$ decreases when solving the same problem~(\figurename~\ref{fig:cd_iteration}A). With less connected topologies, swarms exhibit greater interaction diversity than with more connected ones. Given previous studies, this is an expected result: short topological distances lead to fast information flow, which decreases the diversity\cite{Bratton2007}. Our results revealed that the interaction diversity in the swarm depends on the problem; the same topology leads to distinct levels of diversity when optimizing different functions. Though the topology bounds the interactions among particles, the swarm organizes the information flows to optimize a function. 

Indeed, swarm intelligence systems have the capability to self-organize during the optimization process. To assess the relationship between swarm search and interaction diversity, we examine the pace of $f_\Delta$ at which a swarm improves and the interaction diversity at each iteration. We find that ${\rm ID}$  exhibits a non-trivial relationship with $f_\Delta$, as seen in \figurename~\ref{fig:cd_iteration}B for the function $F_2$. The average $f_\Delta$ increases with the average ${\rm ID}$ until reaches a maximum pace after which $f_\Delta$ decreases with ${\rm ID}$. The increase of diversity in the social interactions of the swarm leads to faster swarm pace only until a certain level of diversity; then the swarm starts to slow down---swarm dynamics that impact the overall swarm performance, as seen in \figurename~\ref{fig:cd_iteration}C. We also find a non-trivial association between $k$-regular topologies and the best fitness found at the end of the optimization process. From global to $30$-regular topologies, the fitness decreases from $8.06\times10^{3}$ and improves down to $6.77\times10^{3}$, then deteriorates up to $1.01\times10^{4}$. 

\subsubsection*{Discussion}
Our results demonstrate the capability of interaction diversity ${\rm ID}$ to explain the behavior of the swarm during the optimization process in the Particle Swarm Optimization technique.  ${\rm ID}$ enables us to identify changes in the way information flows within the swarm regardless of the type of problem. The leverage capability of the proposed approach brings the possibility to identify imbalances during the search process and to further understand the flow of information within the swarm. For example, more than using this approach to select which is the best topology for a particular problem\cite{6855840}, one can propose adaptive mechanisms to adjust the search mode during the search process.

The interaction diversity is a general measure to assess swarm-based systems because it does not consider peculiarities associated with the swarm metaphor. The approach is defined over the structure of the network---the \textit{interaction space}---which is entirely based on the social interactions. This approach can also help researchers to perform parametric analyses; due to the lack of analytic tools, parametric studies tend to consider simplified versions of the algorithm\cite{clerc2002}. 

\section*{Conclusions}

Bees, ants, birds, bats, and many other animals have inspired several swarm-based algorithms, but the literature still fails to explain their differences and their complex behavior---a situation that potentially prevents us from understanding and improving such algorithms. In the field, we often describe the differences between the techniques or their versions via the performance achieved when solving distinct problems. This black-box approach has enabled the area to grow over the years and to develop excellent general-use tools. This approach, however, lacks \emph{interpretability} or {\em explainability}. How to interpret, for instance, that including a diversity procedure improves the performance of a swarm algorithm? Is this modification the same as using a different algorithm? With this opaque approach, we miss the opportunity to understand swarm intelligence.   

The main barrier to understanding the swarm complex behavior is the discontinuity between the micro-level actions of individuals and the macro-level behavior of the swarm. In our work, we argue that the {\it interaction network} is at a meso level that can help to explain and understand these systems. With this approach, we can examine a system via an intermediary structure that emerges from the social interactions within the swarm. We can now analyze the patterns of these self-organized interactions. The interaction network also grants an agnostic representation of swarm systems in the \textit{interaction space} which provides us with a more general perspective of swarm-based algorithms. 

To verify the plausibility of this network-based approach, we considered four different swarm-based algorithms with distinct natural inspirations, and then we focused on one of the most popular optimization technique, namely, the Particle Swarm Optimization. We also discussed the social interaction in other self-organization mechanisms to guide definitions of their interaction network. In the analysis of the four algorithms, we showed that the interaction network provides us the means to study them from a general perspective. In the in-depth analysis of the PSO algorithm, we found that the interaction network helps us to disentangle complex features of swarm systems. We analyzed its interplay with the quality and improvement of the fitness, and we found that some characteristics of the interaction network can be used to explain parametric settings in the algorithm. Specifically, we studied the diversity in the network (i.e., the Interaction Diversity). Our results revealed that different communication topologies lead the swarm to distinct search modes that also depend on the problem landscape.

The network-based perspective of swarms unfolds a pathway to researchers to study these systems comprehensively. This perspective creates opportunities on two fronts. First, it brings the required general viewpoint to build an objective classification of swarm-based algorithms. This classification guides the algorithm selection for problem-solving and the development of novel or hybrid methods. Second, the network empowers scholars to examine swarms from an intermediate level that is important to understand the complex behavior of these systems. At this meso level, we expose the effects of the swarm rules which are hidden in the swarm behavior.

In our research, we developed a general concept for the interaction network. This definition has limitations that might illuminate research directions. For instance, we lack a procedure to identify the most appropriate description for a given algorithm. Also, we use a static network definition (i.e., constant number of nodes) that might be inadequate to model some swarm algorithms, especially the ones with evolutionary operators such as selection. In this study, we limited our analyses to optimization algorithms and only performed numerical analyses on the PSO algorithm. Further efforts are needed to investigate the application of the framework on different types of swarm-based algorithms. Still, here we proposed a general approach that makes possible to perform parametric analyses, quantify differences between methods, balance techniques with hybrid or adaptive versions, and build meso-level mechanisms. These are also directions for future research.

\bibliography{sciadvfile}
\bibliographystyle{unsrt}

%

\section*{Declarations}

\subsection*{Availability of data and materials} 
The code for the analyses performed here can be accessed at \url{http://github.com/macoj/network_swarm/}.

\subsection*{Competing Interests} 
The authors declare that they have no competing financial interests.

\subsection*{Authors' contributions} 
MO and DP conceived and designed the experiments; MO and MM performed the experiments; MO, DP, and MM analyzed the data; All authors wrote the paper and have read and approved the final manuscript.

\subsection*{Funding} 
The authors received no specific funding for this work.

\subsection*{Acknowledgments} 
Not applicable.



\end{document}